\documentclass[11pt,a4paper]{article}
\usepackage[hyperref]{acl2020}
\usepackage{times}
\usepackage{graphicx}
\usepackage{wrapfig}
\usepackage{shortcuts}
\usepackage{dsfont}
\usepackage{latexsym}
\usepackage[linesnumbered,algoruled,boxed,noend]{algorithm2e}
\usepackage{amssymb}
\usepackage{amsmath}
\usepackage{url} 
\SetKwInOut{Parameter}{parameter}
\usepackage{enumitem}
\usepackage{mathtools}
\usepackage{cleveref}
\usepackage{multirow}
\usepackage{hhline}
\usepackage[export]{adjustbox}
\usepackage{booktabs,array}
\usepackage{amsmath, bm}
\usepackage{color, colortbl}
\SetKwInput{KwInput}{Input}
\SetKwInput{KwOutput}{Output}
\newcolumntype{P}[1]{>{\centering\arraybackslash}p{#1}}
\usepackage{booktabs,makecell,tabularx} 
\setitemize{noitemsep,topsep=0pt,parsep=0pt,partopsep=0pt}
\let\oldnl\nl
\definecolor{Gray}{gray}{0.85}
\definecolor{LightCyan}{rgb}{0.88,1,1}
\newcommand{\nonl}{\renewcommand{\nl}{\let\nl\oldnl}}

\aclfinalcopy 
\begin{document}

\title{TriggerNER: Learning with Entity Triggers as Explanations \\for Named Entity Recognition}


\author{
Bill Yuchen Lin\textsuperscript{\dag}\thanks{~~{The first two authors contributed equally.}},~
Dong-Ho Lee\textsuperscript{\dag}$^*$,~
Ming Shen\textsuperscript{\dag},~
Ryan Moreno\textsuperscript{\dag},\\ 
\textbf{{Xiao Huang}\textsuperscript{\dag},~Prashant Shiralkar\textsuperscript{\ddag},~
{Xiang Ren}\textsuperscript{\dag}}\\
\textsuperscript{\dag}{University of Southern California} \quad  \textsuperscript{\ddag} Amazon \\
{\texttt{\{yuchen.lin,dongho.lee,shenming,morenor\}@usc.edu},} \\
\texttt{\{huan183,xiangren\}@usc.edu, shiralp@amazon.com}
}


\maketitle

\begin{abstract}

Training neural models for named entity recognition (NER) in a new domain often requires additional human annotations (e.g., tens of thousands of labeled instances) that are usually expensive and time-consuming to collect. 
Thus, a crucial research question is how to obtain supervision in a cost-effective way.
In this paper, we introduce ``entity triggers,'' 
an effective proxy of human explanations for facilitating label-efficient learning of NER models.
An entity trigger is defined as a group of words in a sentence that helps to explain why humans would recognize an entity in the sentence. 

%
We crowd-sourced 14k entity triggers for two well-studied NER datasets\footnote{We  release the entity triggers and code at \url{http://github.com/INK-USC/TriggerNER}}.
Our proposed model, \textit{Trigger Matching Network},  jointly learns trigger representations and soft matching module with self-attention such that can generalize to unseen sentences easily for tagging. 
Our framework is significantly more cost-effective than the traditional neural NER frameworks. Experiments show that using only 20\% of the trigger-annotated sentences results in a comparable performance as using 70\% of conventional annotated sentences.


\end{abstract}

\section{Introduction}\label{sec:intro}
Named entity recognition (NER) is a fundamental information extraction task that focuses on extracting entities from a given text and classifying them using pre-defined categories (e.g., persons, locations, organizations)~\cite{nadeau2007survey}.
Recent advances in NER have primarily focused on training neural network models with an abundance of human annotations, yielding state-of-the-art results~\cite{DBLP:conf/naacl/LampleBSKD16}.
However, collecting human annotations for NER is expensive and time-consuming, especially in social media messages~\cite{Lin2017MultichannelBM} and technical domains such as biomedical publications, financial documents, legal reports, etc.
As we seek to advance NER into more domains with less human effort, how to learn neural models for NER in a cost-effective way becomes a crucial research problem.

The standard protocol for obtaining an annotated NER dataset involves an annotator selecting token spans in a sentence as mentions of entities, and labeling them with an entity type.
However, such annotation process provides limited supervision \textit{per example}.
Consequently, one would need large amount of annotations in order to train high-performing models for a broad range of entity types, which can clearly be cost-prohibitive. 
The key question is then \textit{how can we learn an effective NER model in presence of limited quantities of labeled data}?

\begin{figure}[t]
\centering
	\includegraphics[width=0.8\linewidth]{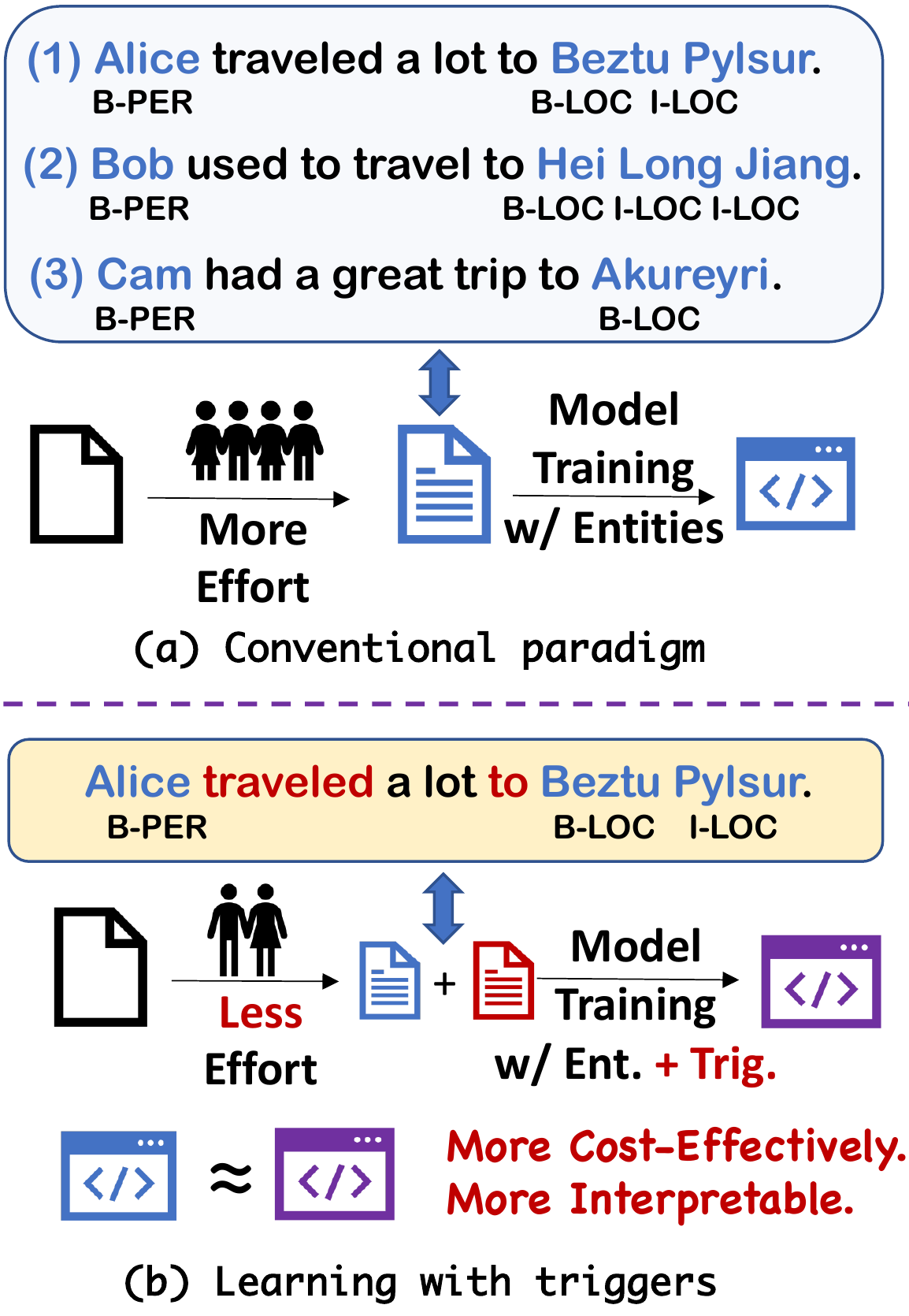}
	\caption{{\textbf{Comparison between (a) conventional learning paradigm and (b) our proposed trigger-based method.} Learning with triggers produces more cost-effective and interpretable NER models.  }}
	\label{fig:intro}
\end{figure}

We, as humans, recognize an entity within a sentence based on certain words or phrases that act as cues. 
For instance, we could infer that \textit{`Kasdfrcxzv'} is likely to be a location entity in the sentence ``\textit{Tom \underline{traveled} a lot last year \underline{in} Kasdfrcxzv.}''
We recognize this entity because of the cue phrase ``\textit{travel ... in},'' which suggests there should be a location entity following the word 'in'.
We call such phrases ``entity triggers.'' Similar to the way these triggers guide our recognition process, we hypothesize that they can also help the model to learn to generalize efficiently, as shown in Fig.~\ref{fig:intro}.

\begin{figure*}[t]
	\centering
	\includegraphics[width=0.85\linewidth]{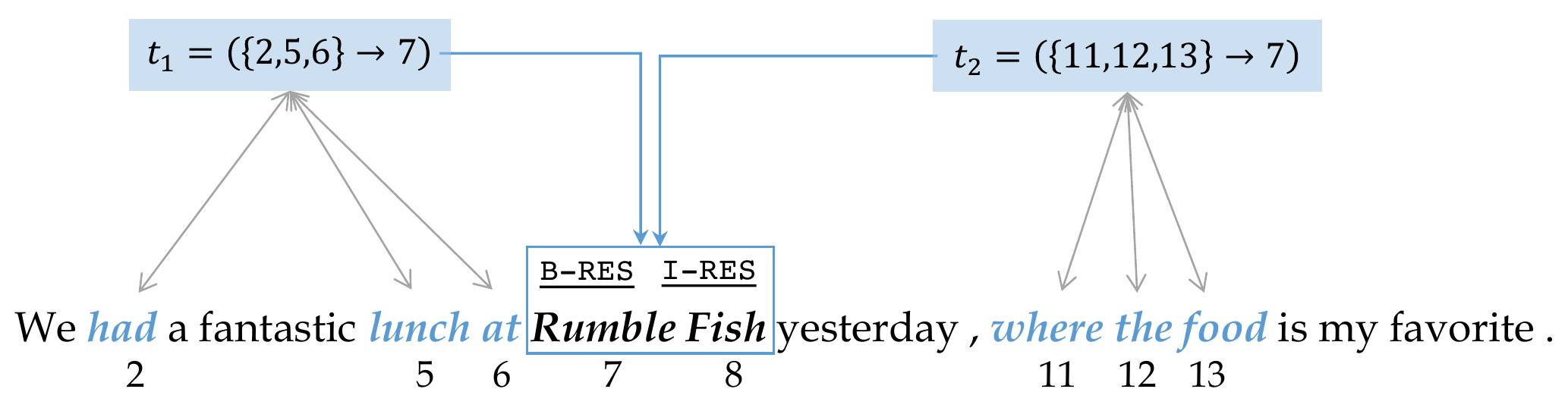}
	\caption{We show two individual \textbf{entity triggers}: $t_1$ (``\textit{had ... lunch at}'') and $t_2$ (``\textit{where the food}''). Both are associated to the same entity mention {\textit{``Rumble Fish''} (starting from 7th token)} typed as  restaurant (\texttt{RES}).}
	\vspace{-10pt}
	\label{fig:trigex}
\end{figure*}

Specifically, we define an \textbf{\textit{``entity trigger''}} (or trigger for simplicity) as a group of words that can help explain the recognition process of a particular entity in the same sentence.
For example, in~\figref{fig:trigex}, \textit{``had ... lunch at''}\footnote{Note that a trigger can be a discontinuous phrase.} and \textit{``where the food''} are two distinct \textit{triggers} associated with the \textsc{Restaurant} entity {\textit{``Rumble Fish.''}}
An entity trigger should be a necessary and sufficient cue for humans to recognize its associated entity even if we mask the entity with a random word.
Thus, unnecessary words such as ``\textit{fantastic}'' should not be considered part of the entity trigger.

In this paper, we argue that a combination of entity triggers and standard entity annotations can enhance the generalization power of NER models.
This approach is more powerful because unlabeled sentences, such as \textit{``Bill \underline{enjoyed a great dinner} with Alice \underline{at} Zcxlbz.''}, can be matched with the existing trigger \textit{``had ... lunch at''} via their semantic relatedness.
This makes it easier for a model to recognize \textit{``Zcxlbz''} as a \textsc{Restaurant} entity.
In contrast, if we only have the entity annotation itself (i.e., \textit{``Rumble Fish''}) as supervision, the model will require many similar examples in order to learn this simple pattern.
Annotation of triggers, in addition to entities, does not incur significantly additional effort because the triggers are typically short, and more importantly, the annotator has already comprehended the sentence, identifying their entities as required in the traditional annotation. 
On the benchmark datasets we consider, the average length of a trigger in our crowd-sourced dataset is only 1.5-2 words.
Thus, we hypothesize that using triggers as additional supervision is a more cost-effective way to train models.

We crowd-sourced annotations of 14,708 triggers on two well-studied NER datasets to study their usefulness for the NER task.
We propose a novel framework named Trigger Matching Network that learns trigger representations indicative of entity types during the training phase, and identifies triggers in an unlabeled sentence at inference time to guide a traditional entity tagger for delivering better overall NER performance. Our TMN framework consists of three components: 1) a trigger encoder to learn meaningful trigger representations for an entity type, 2) a semantic trigger matching module for identifying triggers in a new sentence, and 3) an entity tagger that leverages trigger representations for entity recognition (as present in existing NER frameworks).
Different from conventional training, our learning process consists of two stages, in which the first stage comprises jointly training a trigger classifier and the semantic trigger matcher, followed by a second stage that leverages the trigger representation and the encoding of the given sentence using an attention mechanism to learn a sequence tagger.

Our contributions in this paper are as follows:
\begin{itemize}
    \item We introduce the concept of ``entity triggers,'' a novel form of explanatory annotation for named entity recognition problems.  We crowd-source and publicly release 14k annotated entity triggers on two popular datasets: \textit{CoNLL03} (generic domain), \textit{BC5CDR} (biomedical domain).
    \item We propose a novel learning framework, named \textit{Trigger Matching Network}, which encodes entity triggers and softly grounds them on unlabeled sentences to increase the effectiveness of the base entity tagger (Section~\ref{sec:tmn}). 
    \item Experimental results (Section~\ref{sec:exp}) show that the proposed trigger-based framework is significantly more cost-effective. The TMN uses 20\% of the trigger-annotated sentences from the original CoNLL03 dataset, while achieving a comparable performance to the conventional model using 70\% of the annotated sentences. 
\end{itemize}


\section{Problem Formulation}
\label{sec:problem}

We consider the problem of \textit{how to cost-effectively learn a model for NER using entity triggers}. In this section, we introduce basic concepts and their notations, present the conventional data annotation process for NER, and provide a formal task definition for learning using entity triggers.

In the conventional setup for supervised learning for NER, we let $\mathbf{x}=[x^{(1)}, x^{(2)}, \cdots, x^{(n)}]$ denote a sentence in the labeled training corpus $\mathcal{D}_{L}$.
Each labeled sentence has a NER-tag sequence $\textbf{y}=[y^{(1)}, y^{(2)}, \cdots, y^{(n)}]$, where $y^{(i)}\in \mathcal{Y}$ and  $\mathcal{Y}$ can be \{\texttt{O}, \texttt{B-PER}, \texttt{I-PER}, \texttt{B-LOC}, \texttt{I-LOC}, $\cdots$\}. The possible tags come from a BIO or BIOES tagging schema for segmenting and typing entity tokens.
Thus, we have $\mathcal{D}_{L}=\{(\mathbf{x_i}, \mathbf{y_i})\}$, and an unlabeled corpus $\mathcal{D}_{U}=\{\mathbf{x_i}\}$.

We propose to annotate entity triggers in sentences.
We use $T(\mathbf{x},\mathbf{y})$ to represent the set of annotated entity triggers, where each trigger $t_i\in T(\mathbf{x},\mathbf{y})$ is associated with an entity index $e$ and a set of word indices $\{w_i\}$.
Note that we use the index of the first word of an entity as its entity index.
That is, $t = (\{w_1, w_2, \cdots\}\rightarrow{e})$, where $e$ and $w_i$ are integers in the range of $[1,|\mathbf{x}|]$. \quad
For instance, in the example shown in Figure~\ref{fig:trigex}, the trigger \textit{``had ... lunch at''} can be represented as a trigger $t_1=(\{2,5,6\}\rightarrow{7})$, because this trigger specifies the entity starting at index $7$, \textit{``Rumble''}, and it contains a set of words with indices: \textit{``had''} ($2$), \textit{``lunch''} ($5$), and \textit{``at''} ($6$).
Similarly, we can represent the second trigger \textit{``where the food''} as $t_2 = (\{11,12,13\}\rightarrow{7})$.
Thus, we have $T(\mathbf{x},\mathbf{y})=\{t_1, t_2\}$ for this sentence.

Adding triggers creates a new form of data $\mathcal{D}_{T} = \{(\mathbf{x_i},\mathbf{y_i}, T(\mathbf{x_i},\mathbf{y_i})\}$. 
Our goal is to learn a model for NER from a trigger-labeled dataset $\mathcal{D}_T$, such that we can achieve comparable learning performance to a model with a much larger $\mathcal{D}_L$.
 
\section{{Trigger Matching Networks}}
\label{sec:method}
\label{sec:tmn}

We now present our framework for a more cost-effective learning method for NER using triggers. We assume that we have collected entity triggers (the trigger collection process is discussed in Section~\ref{sec:trigger}).
At a high-level, we aim to learn trigger representations for entity types that allow the entity tagger to generalize for sentences beyond the training phase. Our intuition is that triggers acting as cues for the same named-entity type should have similar trigger representations, and thus triggers can be identified in an unlabeled sentence at inference time by soft-matching between the sentence representation and trigger representations seen during training. 
We perform such soft-matching using a  self-attention mechanism.

\begin{figure*}[t]
 	\centering 
	\includegraphics[width=0.95\linewidth]{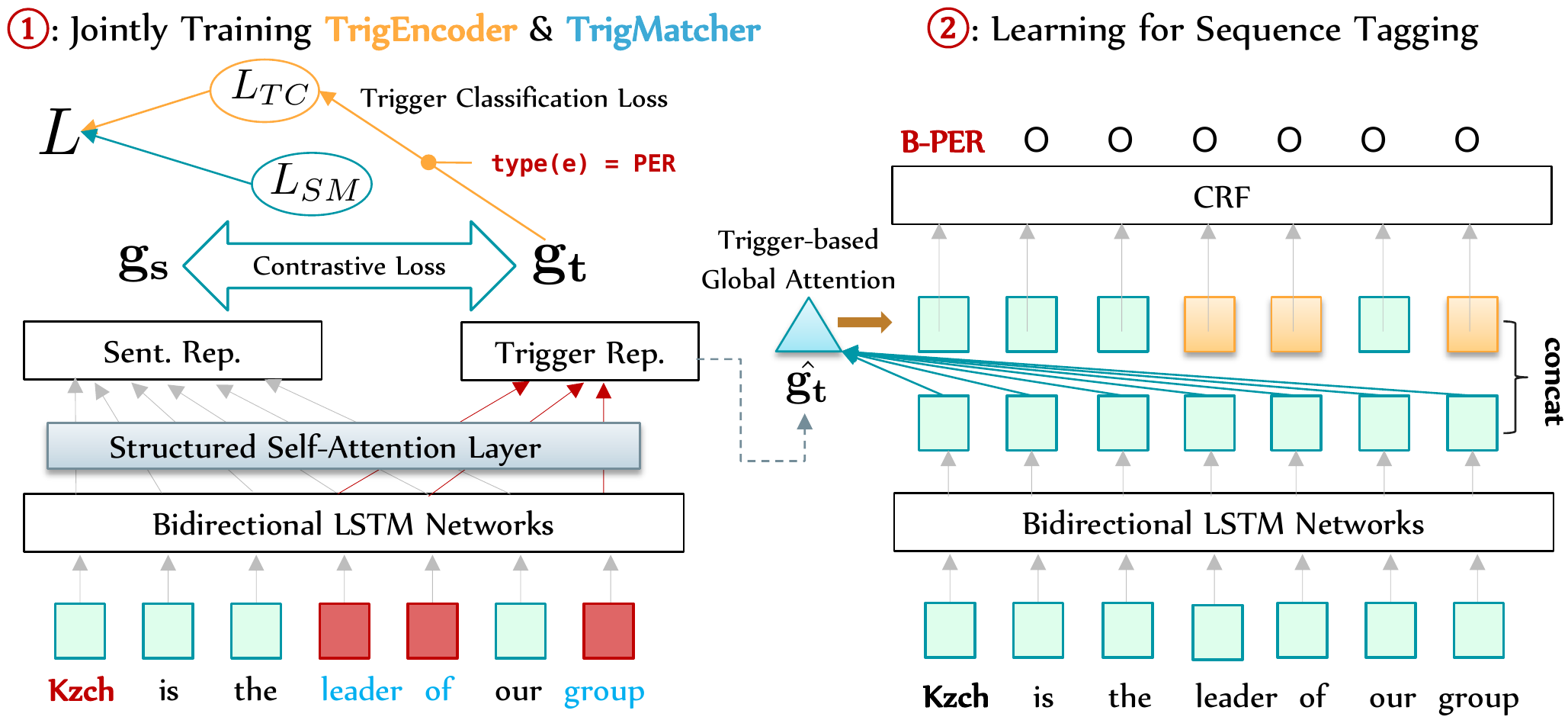}
	\caption{\textbf{Two-stage training of the \textit{Trigger Matching Network}.} We first jointly train the {\texttt{TrigEncoder}} (via trigger classification) and the \texttt{TrigMatcher} (via contrastive loss). Then, we reuse the training data trigger vectors as attention queries in the \texttt{SeqTagger}.} 
	\label{fig:framework}
\end{figure*}

We propose a straightforward yet effective framework, named \textit{Trigger Matching Networks} (TMN), consisting of a trigger encoder ({\texttt{TrigEncoder}}), a semantic-based trigger matching module (\texttt{TrigMatcher}), and a base {sequence tagger} (\texttt{SeqTagger}). 
We have two learning stages for the framework: the first stage (Section~\ref{sec:firststage}) jointly learns the \texttt{TrigEncoder} and \texttt{TrigMatcher}, and the second stage (Section~\ref{sec:secondstage}) uses the trigger vectors to learn NER tag labels.
Figure~\ref{fig:framework} shows this pipeline. We introduce the inference in Section~\ref{sec:inference}.



\subsection{Trigger Encoding \& Semantic Matching}
\label{sec:firststage}


Learning trigger representations and semantically matching them with sentences are inseparable tasks.
Desired trigger vectors capture the semantics in a shared embedding space with token hidden states, such that sentences and triggers can be semantically matched.
Recall the example we discussed in Sec.~\ref{sec:intro}, \textit{``enjoyed a great dinner at''} versus \textit{``had ... lunch at.''}
Learning an attention-based matching module between entity triggers and sentences is necessary so that triggers and sentences can be semantically matched.
Therefore, in the first stage, we propose to jointly train the trigger encoder  (\texttt{TrigEncoder}) and the attention-based trigger matching module  (\texttt{TrigMatcher}) using a shared embedding space.

Specifically,
for a sentence $\mathbf{x}$ with multiple entities $\{e_1, e_2,\cdots\}$, for each entity $e_i$ we assume that there is a set of triggers $T_i=\{t^{(i)}_1, t^{(i)}_2, \cdots\}$ without loss of generality.
To enable more efficient batch-based training, we reformat the trigger-based annotated dataset $\mathcal{D}_{T}$ such that each new sequence contains only one entity and one trigger.
We then create a training instance by pairing each entity with one of its triggers, denoted $(\mathbf{x}, e_i, t^{(i)}_j)$.

For each reformed training instance $(\mathbf{x}, e, t)$, we first apply a bidirectional LSTM (BLSTM) on the sequence of word vectors\footnote{Here, by ``word vectors'' we mean the concatenation of external GloVe~\cite{Pennington2014GloveGV} word embeddings and char-level word representations from  a trainable CNN network~\cite{DBLP:conf/acl/MaH16}. } of $\mathbf{x}$, obtaining a sequence of hidden states that are the contextualized word representations $\mathbf{h}_i$ for each token $x_i$ in the sentence. 
We use $\mathbf{H}$ to denote the matrix containing the hidden vectors of all of the tokens, and we use $\mathbf{Z}$ to denote the matrix containing the hidden vectors of all  trigger tokens inside the trigger $t$.

In order to learn an attention-based representation of both triggers and sentences, we follow the self-attention method introduced by ~\cite{selfattentive} as follows:
{
{
		\begin{align*} 
			\vec{a}_{sent}&=\operatorname{SoftMax}\left(W_{2} \tanh \left(W_{1} \mathbf{H}^{T}\right)\right)\\
			\mathbf{g_s}&=\vec{a}_{sent}\mathbf{H}\\
			\vec{a}_{trig}&=\operatorname{SoftMax}\left(W_{2} \tanh \left(W_{1} \mathbf{Z}^{T}\right)\right)\\
			\mathbf{g_t}&=\vec{a}_{trig}\mathbf{Z}
		\end{align*} 
	}
}  
$W_1$ and $W_2$ are two trainable parameters for computing self-attention score vectors $\vec{a}_{sent}$ and $\vec{a}_{trig}$.
We obtain a vector representing the weighted sum of the token vectors in the entire sentence as the final sentence vector $\mathbf{g_s}$. Similarly, $\mathbf{g_t}$ is the final trigger vector, representing the weighted sum of the token vectors in the trigger.

We want to use the type of the associated entity as supervision to guide the trigger representation.
Thus, the trigger vector $\mathbf{g_t}$ is further fed into a multi-class classifier to predict the \textit{type} of the associated entity $e$ (such as \texttt{PER}, \texttt{LOC}, etc) which we use $\texttt{type}(e)$ to denote.
The loss of the trigger classification is as follows:$$L_{TC}=-\sum \log \mathrm{P}\left(\texttt{type}(e)~|~ \mathbf{g_t}; \theta_{TC}\right),$$
where $\theta_{TC}$ is a model parameter to learn.

Towards learning to match triggers and sentences based on attention-based representations, we use contrastive loss~\cite{hadsell2006dimensionality}.
The intuition is that similar triggers and sentences should have close representations (i.e., have a small distance between them, $d$).
We create negative examples (i.e., mismatches) for training by randomly mixing the triggers and sentences,
because \texttt{TrigMatcher} needs to be trained with both positive and negative examples of the form (sentence, trigger, label). 
For the negative examples, we expect a margin $m$ between their embeddings. 
The contrastive loss of  the soft matching is defined as follows, where $\mathds{1}_{\text{matched}}$ is $1$ if the trigger was originally in this sentence and $0$ if they are not:
{
	{
		\begin{gather*}  
		d=\left\|\mathbf{g_s}-\mathbf{g_t}\right\|_{2}\\
			\hspace{-50pt} L_{SM} = \mathds{1}_{\text{matched}} \frac{1}{2}\left(d\right)^{2} + \\ \hspace{20pt} (1-\mathds{1}_{\text{matched}}) \frac{1}{2}\left\{\max \left(0, m-d\right)\right\}^{2}
		\end{gather*} 
	}
}  
The joint loss of the first stage is thus $L = L_{TC} + \lambda L_{SM}$, where $\lambda$ is a hyper-parameter to tune.
\begin{figure*}[t]
 	\centering 
	\includegraphics[width=0.95\linewidth]{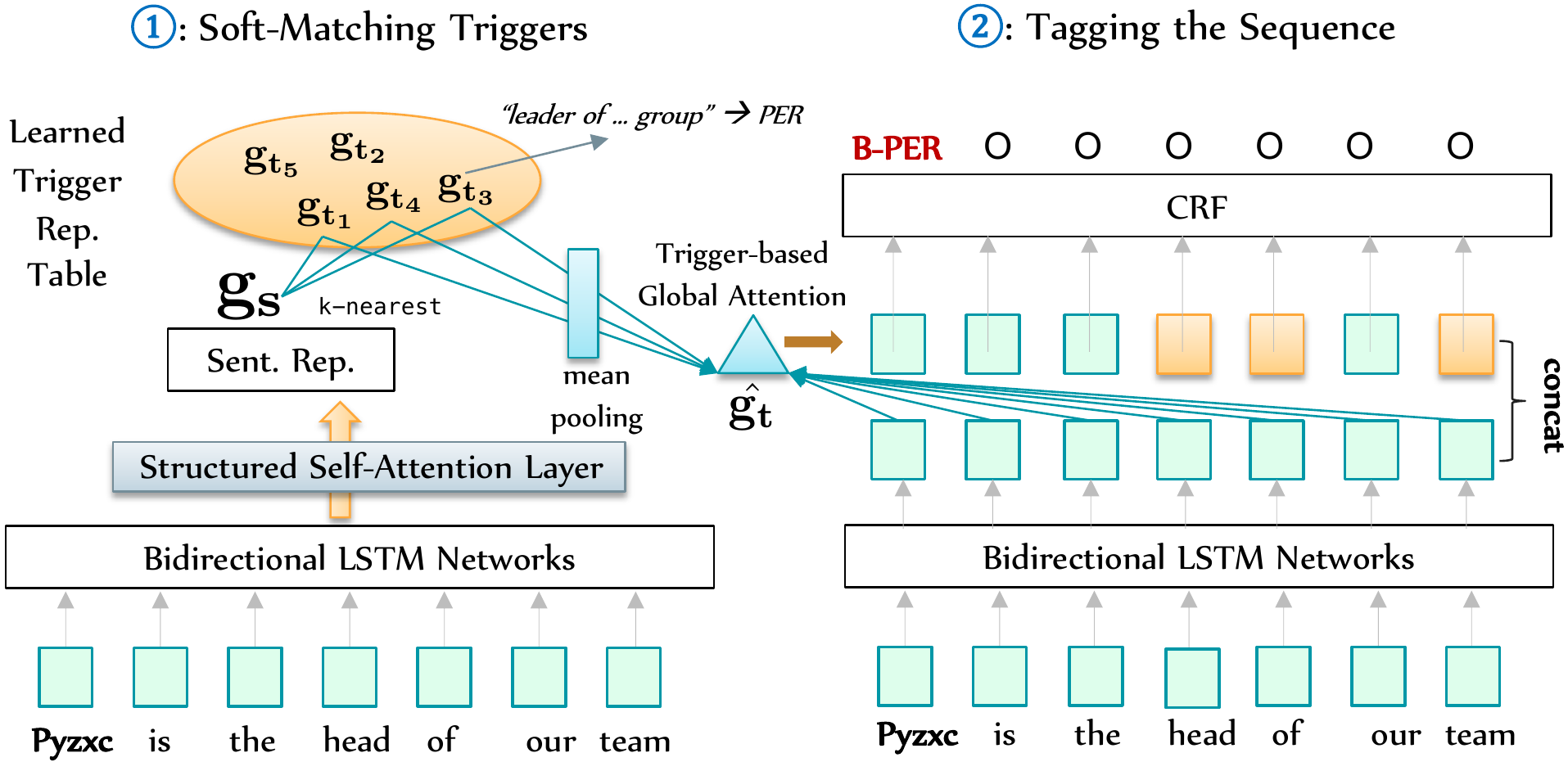}
	\caption{\textbf{The {inference} process of the TMN framework.} It uses the \texttt{TrigMatcher} to retrieve the $k$ nearest triggers and average their trigger vectors as the attention query for the trained \texttt{SeqTagger}. Thus, an unseen cue phrase (e.g., \textit{``head of ... team''}) can be matched with a seen trigger (e.g., \textit{``leader of ... group''}).} 
	\label{fig:inference}
\end{figure*}
\subsection{Trigger-Enhanced Sequence Tagging}
\label{sec:secondstage}
The learning objective in this stage is to output the tag sequence $\mathbf{y}$.
Following the most common design of neural NER architecture, BLSTM-CRF~\cite{DBLP:conf/acl/MaH16}, we incorporate the entity triggers as attention queries to train a trigger-enhanced sequence tagger for NER. 
Note that the BLSTM used in the the \texttt{TrigEncoder} and \texttt{TrigMatcher} modules is the same BLSTM we use in the \texttt{SeqTagger} to obtain $\mathbf{H}$, the matrix containing the hidden vectors of all of the tokens.
Given a sentence $\mathbf{x}$, we use the previously trained \texttt{TrigMatcher} to compute the mean of all the trigger vectors $\hat{\mathbf{g_t}}$ associated with this sentence.
Following the conventional attention method~\cite{luong2015effective}, 
we incorporate the mean trigger vector as the query, creating a sequence of attention-based token representations, $\mathbf{H}'$.
\begin{gather*} 
\vec{\alpha}  = \operatorname{SoftMax}\left(\boldsymbol{v}^{\top} \tanh \left({U}_{1}\mathbf{H}^T + {U}_{2}\hat{\mathbf{g_t}}^T \right)^{\top}\right)\\
\mathbf{H'} =  \vec{\alpha}~\mathbf{H}
\end{gather*}  
$U_1$, $U_2$, and $v$ are trainable parameters for computing the trigger-enhanced attention scores for each token.
Finally, we concatenate the original token representation $\mathbf{H}$ with the trigger-enhanced one $\mathbf{H}'$ as the input ($[\mathbf{H};\mathbf{H}']$) to the final CRF tagger.
Note that in this stage, our learning objective is the same as conventional NER, which is to correctly predict the tag for each token.

\subsection{Inference on Unlabeled Sentences}
\label{sec:inference}

When inferencing tags on unlabeled sentences,
we do not know the sentence's triggers.
Instead, we use the \texttt{TrigMatcher} to compute the similarities between the self-attended sentence representations and the trigger representations, using the most suitable triggers as additional inputs to the \texttt{SeqTagger}.
Specifically, we have a trigger dictionary from our training data, $\mathcal{T}=\{t | (\cdot, \cdot, t) \in \mathcal{D}_T\}$.
Recall that we have learned a trigger vector for each of them, and we can load these trigger vectors as a look-up table in memory.
For each unlabeled sentence $\mathbf{x}$, we first compute its self-attended vector $\mathbf{g_s}$ as we do when training the \texttt{TrigMatcher}.
Using L2-norm distances to compute the contrastive loss, we efficiently retrieve the most similar triggers in the shared embedding space of the sentence and trigger vectors.

Then, we calculate $\hat{\mathbf{g_t}}$, the mean of the top $k$ nearest semantically matched triggers, as this serves a proxy to triggers mentioned for the entity type in the labeled data.
We then use it as the attention query for \texttt{SeqTagger}, similarly in Sec.~\ref{sec:secondstage}.
Now, we can produce trigger-enhanced sequence predictions on unlabeled data, as shown in Fig.~\ref{fig:inference}.

\section{Experiments}\label{sec:exp}
\begin{table*}[t]
	\centering
	\scalebox{0.76 
	}{
		\begin{tabular}{ccccccccc}
			\toprule   
			\textbf{Dataset} &      
			\multicolumn{1}{c}{Entity Type}  &  
			\multicolumn{1}{c}{\# of Entities} & 
			\multicolumn{1}{c}{\# of Triggers} & 
			\multicolumn{1}{c}{Avg. \#  of Triggers per Entity} & 
			\multicolumn{1}{c}{Avg. Trigger Length} 
			\\ 
			\midrule
			\textsc{CONLL 2003} & \textsc{PER} & 1,608 & 3,445 & 2.14 & 1.41 \\
			& \textsc{ORG} & 958 & 1,970 & 2.05 & 1.46 \\  
			& \textsc{MISC} & 787 & 2,057 & 2.61 & 1.4 \\ 
		    & \textsc{LOC} & 1,781 & 3,456 & 1.94 & 1.44 \\ 
		    \midrule
			& \textbf{Total} & 5,134 & 10,938 & 2.13 & 1.43\\ 
			\midrule\midrule
			\textsc{BC5CDR} & \textsc{Disease} & 906 & 2,130 & 2.35 & 2.00  \\
			& \textsc{Chemical} & 1,085 & 1,640 & 1.51 & 1.99  \\ 
			\midrule
			& \textbf{Total} & 1,991 & 3,770 & 1.89 & 2.00 \\ 
			\bottomrule
		\end{tabular}
	} 
	\caption{{\textbf{Statistics of the crowd-sourced entity triggers}}.}
	\label{tab:numtrig}
\end{table*}
In this section, we first discuss how to collect entity triggers, and empirically study the data-efficiency of our proposed framework.
\subsection{Annotating Entity Triggers as Explanatory Supervision}
\label{sec:trigger}
We use a general domain dataset CoNLL2003~\cite{conll} and a bio-medical domain dataset BC5CDR~\cite{bc5cdr}.
Both datasets are well-studied and popular in evaluating the performance of neural named entity recognition models such as BLSTM-CRF~\cite{DBLP:conf/acl/MaH16}.

In order to collect the entity triggers from human annotators, 
we use \textit{Amazon SageMaker Ground Truth}\footnote{An advanced version of \textit{Amazon Mechanical Turk}. \url{https://aws.amazon.com/sagemaker/}} to crowd-source entity triggers.
More recently, \citeauthor{LEANLIFE} (\citeyear{LEANLIFE}) developed an annotation framework, named LEAN-LIFE, which supports our proposed trigger annotating.
Specifically, we sample 20\% of each training set as our inputs, and then reform them to be the same format as we discussed in Section~\ref{sec:problem}.
Annotators are asked to annotate a group of words that would be helpful in typing and/or detecting the occurrence of a particular entity in the sentence.
We masked the entity tokens with their types so that human annotators are more focused on the non-entity words in the sentence when considering the triggers.
We consolidate multiple triggers for each entity by taking the intersection of the three annotators' results. 
Statistics of the final curated triggers are summarized in~\tabref{tab:numtrig}.
We release the 14k  triggers to the community for future research in trigger-enhanced NER.





\begin{table*}[t]
	\centering
	\scalebox{0.85}
	{
		\begin{tabular}{>{\columncolor{LightCyan}}c|ccc|>{\columncolor{LightCyan}}c|ccc|cccc }
			  \toprule
			\multicolumn{11}{c}{\textbf{CONLL 2003}}\\
			\toprule
			\rowcolor{white} & 
			\multicolumn{3}{c|}{\textsc{BLSTM-CRF}}  & 
			&
			\multicolumn{3}{c|}{\textsc{TMN}}  &  
			\multicolumn{3}{c}{\textsc{TMN + self-training}}  
			\\
			\midrule
			\rowcolor{white}\textbf{sent.}  &
			\multicolumn{1}{c}{Precision}  &    
			\multicolumn{1}{c}{Recall}  &
			\multicolumn{1}{c|}{F1}  &  
			\textbf{trig. } &
			\multicolumn{1}{c}{Precision}  &    
			\multicolumn{1}{c}{Recall}  &
			\multicolumn{1}{c|}{F1} &
			\multicolumn{1}{c}{Precision}  &    
			\multicolumn{1}{c}{Recall}  &
			\multicolumn{1}{c}{F1}  & 
			\\ 
			\midrule 
			5\%  &  70.85 & 67.32 & 69.04 & 3\%  & 76.36 & 74.33 & 75.33 & 80.36 & 75.18 & 77.68 \\
			10\% &  76.57 & 77.09 & 76.83 & 5\%  & 81.28 & 79.16 & 80.2  & 81.96 & 81.18 & 81.57   \\
            20\% &  82.17 & 80.35 & 81.3  & 7\%  & 82.93 & 81.13 & 82.02 & 82.92 & 81.94 & 82.43   \\
            30\% &  83.71 & 82.76 & 83.23 & 10\% & 84.47 & 82.61 & 83.53 & 84.47 & 82.61 & 83.53   \\
            40\% &  85.31 & 83.1  & 84.18 & 13\% & 84.76 & 83.69 & 84.22 & 84.64 & 84.01 & 84.33   \\
            50\% &  85.07 & 83.49 & 84.27 & 15\% & 85.61 & 84.45 & 85.03 & 86.53 & 84.26 & 85.38   \\
            60\% &  85.58 & 84.54 & 85.24 & 17\% & 85.25 & 85.46 & 85.36 & 86.42 & 84.63 & 85.52   \\
            \textbf{\underline{70\%}} &  86.87 & 85.3  & \textbf{\underline{86.08}}&\textbf{ 20\% }& 86.04 & 85.98 & \textbf{86.01} & 87.09 & 85.91 & \textbf{86.5 }  \\
			\midrule
			\multicolumn{11}{c}{\textbf{BC5CDR}}\\
			\toprule
			\rowcolor{white} & 
			\multicolumn{3}{c|}{\textsc{BLSTM-CRF}}  & 
			&
			\multicolumn{3}{c|}{\textsc{TMN}}  &  
			\multicolumn{3}{c}{\textsc{TMN + self-training}}  
			\\
			\midrule
			\rowcolor{white}\textbf{sent.}  &
			\multicolumn{1}{c}{Precision}  &    
			\multicolumn{1}{c}{Recall}  &
			\multicolumn{1}{c|}{F1}  &  
			\textbf{trig.}  &
			\multicolumn{1}{c}{Precision}  &    
			\multicolumn{1}{c}{Recall}  &
			\multicolumn{1}{c|}{F1} &
			\multicolumn{1}{c}{Precision}  &    
			\multicolumn{1}{c}{Recall}  &
			\multicolumn{1}{c}{F1}  & 
			\\ 
			\midrule 
			5\%  &  63.37 & 43.23 & 51.39 & 3\%  & 66.47 & 57.11 & 61.44 & 65.23 & 59.18 & 62.06 \\
			10\% &  68.83 & 60.37 & 64.32 & 5\%  & 69.17 & 73.31 & 66.11 & 68.02 & 66.76 & 67.38   \\
            20\% &  79.09 & 62.66 & 69.92 & 7\%  & 64.81 & 69.82 & 67.22 & 69.87 & 66.03 & 67.9   \\
            30\% &  80.13 & 65.3 & 71.87 & 10\% & 71.89 & 69.57 & 70.71 & 69.75 & 72.75 & 71.22   \\
            40\% &  82.05 & 65.5 & 72.71 & 13\% & 73.36 & 70.44 & 71.87 & 75.11 & 69.31 & 72.1   \\
            \textbf{\underline{50\%} }&  82.56 & 66.58 & \textbf{\underline{73.71}} & 15\% & 70.91 & 72.89 & 71.89 & 71.23 & 73.31 & 72.26  \\
            60\% &  81.73 & 70.74 & 75.84 & 17\% & 75.67 & 70.6 & 73.05 & 77.47 & 70.47 & 73.97   \\
            70\% &  81.16 & 75.29 & 76.12 & \textbf{20\% }& 77.47 & 70.47 & \textbf{73.97} & 75.23 & 73.83 & \textbf{74.52}   \\
			\midrule
			\bottomrule
		\end{tabular}
		
	}
	\caption{\textbf{Labor-efficiency study on BLSTM-CRF and TMN.} ``sent.'' means the percentage of the sentences (labeled only with entity tags) we use for BLSTM-CRF, while ``trig.'' denotes the percentage of the sentences (labeled with both entity tags and trigger tags) we use for TMN. }
	\label{tab:results}
\end{table*}

\subsection{Base model}
We require a base model to compare with our proposed TMN model in order to validate whether the TMN model effectively uses triggers to improve model performance in a limited label setting.
We choose the CNN-BLSTM-CRF~\cite{DBLP:conf/acl/MaH16} as our base model for its wide usage in research of neural NER models and applications.
Our TMNs are implemented within the same codebase and use the same external word vectors from GloVE~\cite{Pennington2014GloveGV}. The hyper-parameters of the CNNs, BLSTMs, and CRFs are also the same.
This ensures a fair comparison between a typical non-trigger NER model and our trigger-enhanced framework.


\subsection{Results and analysis}

\noindent
\textbf{Labeled data efficiency.}
We first seek to study the cost-effectiveness of using triggers as an additional source of supervision. Accordingly, we explore the performance of our model and the baseline for different fractions of the training data. The results on the two datasets are shown in Table 2. We can see that by using only 20\% of the trigger-annotated data, TMN model delivers comparable performance as the baseline model using 50-70\% traditional training data. The drastic improvement in the model performance obtained using triggers thus justifies the slightly additional cost incurred in annotating triggers.

\noindent
\textbf{Self-training with triggers.}
We also do a preliminary investigation of adopting self-training~\cite{Rosenberg2005SemiSupervisedSO} with triggers.
We make inferences on unlabeled data and take the predictions with high confidences as the weak training examples for continually training the model.
The confidence is computed following the MNLP metric~\cite{shen2018deep}, and we take top 20\% every epoch.
With the self-training method, we further improve the TMN model's F-1 scores by about 0.5$\sim$1.0\%.

\begin{figure}[t]
\centering
	\includegraphics[width=0.95\linewidth]{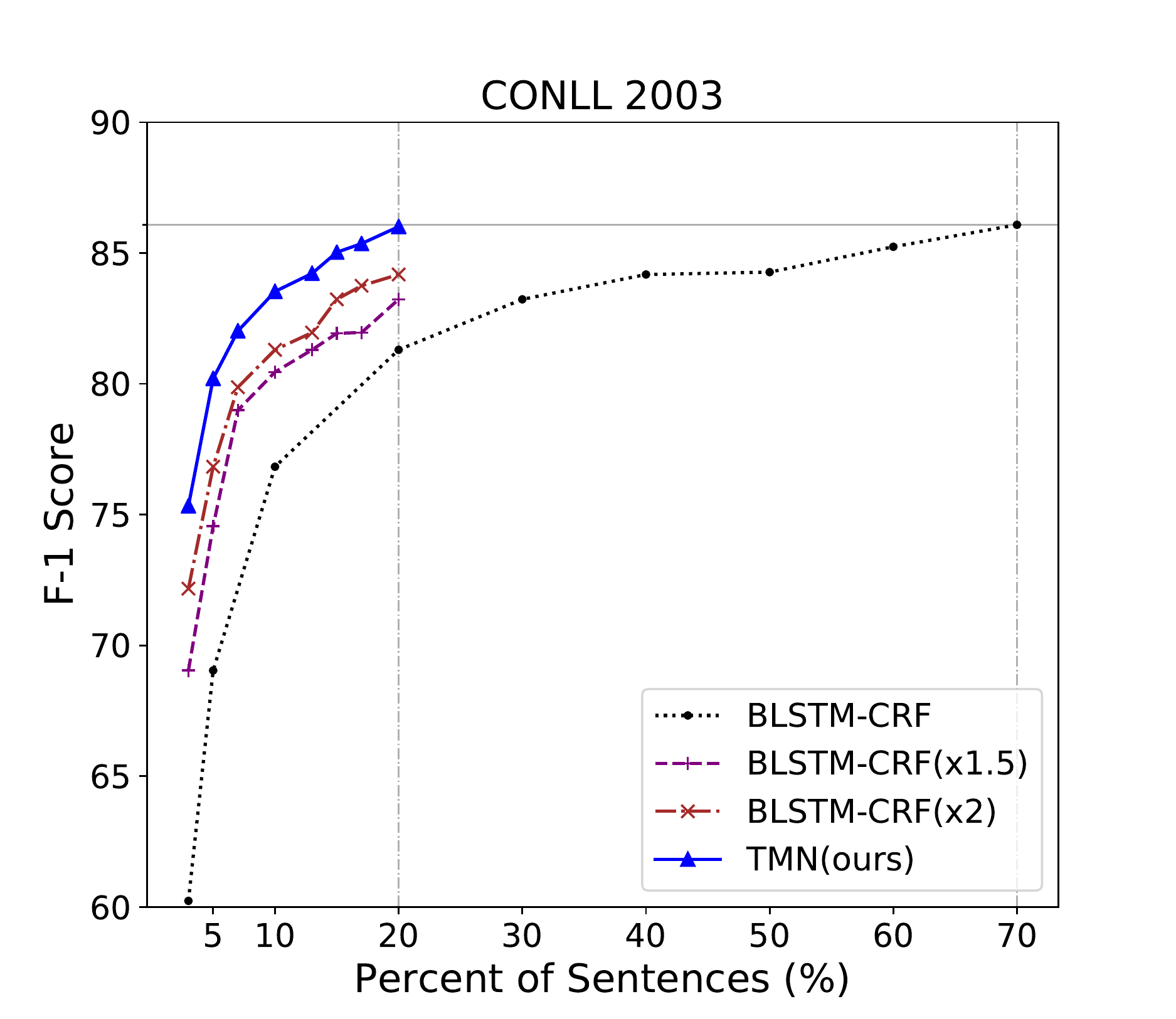}
	\caption{{\textbf{The cost-effectiveness study.} We stretch the curve of BLSTM-CRF parallel to the x-axis by 1.5/2. Even if we assume annotating  entity triggers cost 150/200\% the amount of human effort as annotating entities only, TMN is still much more effective. }}
	\label{fig:curve}
\end{figure}

\noindent
\textbf{Annotation time vs. performance.}
Although it is hard to accurately study the time cost on the crowd-sourcing platform we use\footnote{Annotators may suspend jobs and resume them without interaction with the crowd-sourcing platform.}, based on our offline simulation we argue that annotating both triggers and entities are about $1.5$ times (``BLSTM-CRF (x1.5)'') longer than only annotating entities. our offline simulation.
In \figref{fig:curve}, The x-axis for BLSTM-CRF means the number of sentences annotated with only entities, while for TMN means the number of sentences tagged with both entities and triggers. 
In order to reflect human annotators spending 1.5 to 2 times as long annotating triggers and entities as they spend annotating only entities, we stretch the x-axis for BLSTM-CRF. For example, the line labeled (``BLSTM-CRF (x2)'') associates the actual F1 score for the model trained on 40\% of the sentences with the x-axis value of 20\%.
We can clearly see that the proposed TMN outperforms the BLSTM-CRF model by a large margin.
Even if we consider the extreme case that tagging triggers requires twice the human effort (``BLSTM-CRF (x2)''), the TMN is still significantly more labor-efficient in terms of F1 scores.

\begin{figure*}[t]
	\centering 
	\includegraphics[width=0.9\linewidth]{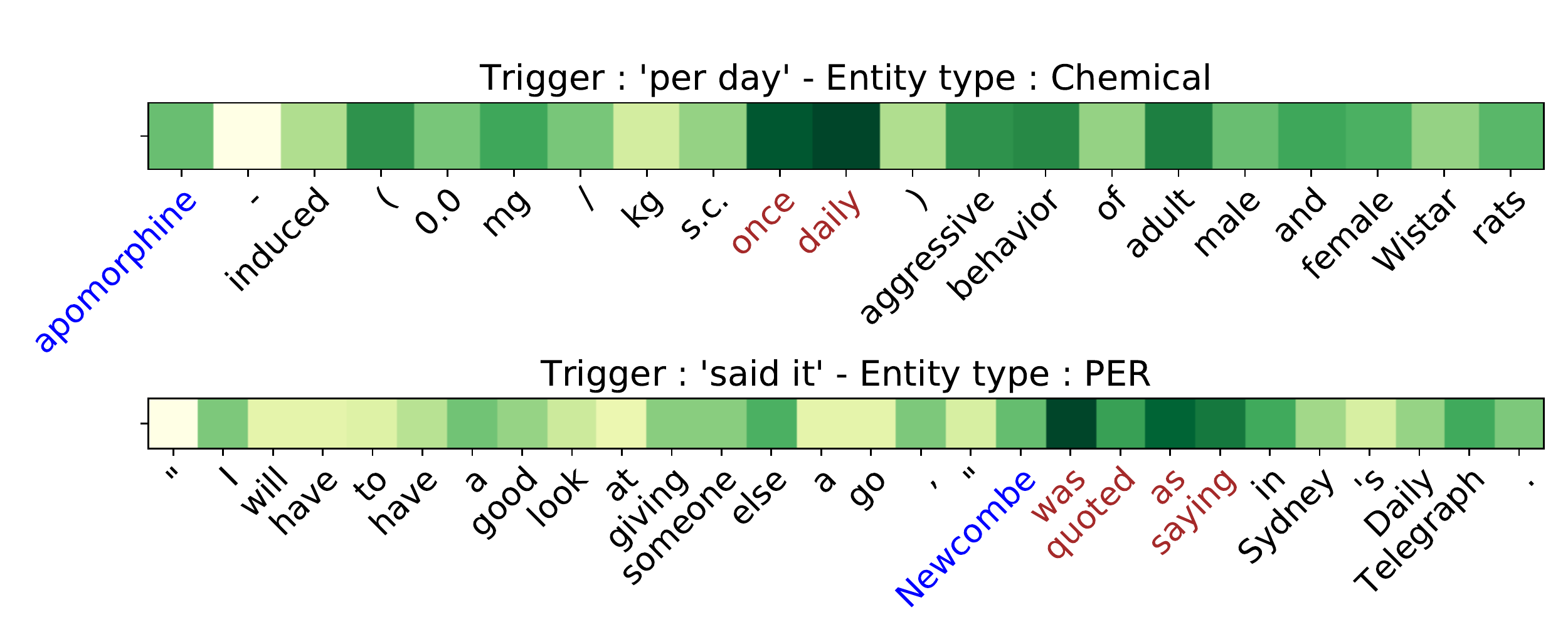}
	\caption{\textbf{Two case studies of trigger attention during inference.} The darker cells have higher attention weights.}
	\label{fig:casestudy}
\end{figure*}

\noindent
\textbf{Interpretability.}
\figref{fig:casestudy} shows two examples illustrating that the trigger attention scores help the TMN model recognize entities. The training data has `per day' as a trigger phrase for chemical-type entities, and this trigger matches the phrase `once daily' in an unseen sentence during the inference phase of \texttt{TrigMatcher}.
Similarly, in CoNLL03 the training data trigger phrase `said it' matches with the phrase `was quoted as saying' in an unlabeled sentence.
These results not only support our argument that trigger-enhanced models such as TMN can effectively learn, but they also demonstrate that trigger-enhanced models can provide reasonable interpretation, something that lacks in other neural NER models.

\section{Related Work}\label{sec:rel_work}


Towards low-resource learning for NER,
recent works have mainly focused on dictionary-based distantly supervision~\cite{autoner, yangner, Liu2019TowardsIN}.
These approaches create an external large dictionary of entities, and then regard hard-matched sentences as additional, noisy-labeled data for learning a NER model.
Although these approaches largely reduce human efforts in annotating, the quality of matched sentences is highly dependent on the coverage of the dictionary and the quality of the corpus.
The learned models tend to have a bias towards entities with similar surface forms as the ones in dictionary. Without further tuning under better supervision, these models have low recall~\cite{Cao2019LowResourceNT}.
\textit{Linking rules}~\cite{safranchik:aaai20} focuses on the 
votes on whether adjacent elements in the sequence belong
to the same class. 
Unlike these works aiming to get rid of training data or human annotations, our work focuses on how to more cost-effectively utilize human efforts.

Another line of research which also aims to use human efforts more cost-effectively is active learning~\cite{shen2018deep,Lin2019AlpacaTagAA}.
This approach focuses on instance sampling and the human annotation UI, asking workers to annotate the most useful instances first.
However, a recent study~\cite{Lipton2018PracticalOT} argues that actively annotated data barely helps when training new models. 
Transfer learning approaches ~\cite{Lin2018NeuralAL} and aggregating multi-source supervision~\cite{Lan2020} are also studied for using less expensive supervision for NER, while these methods usually lack clear rationales to advise annotation process unlike the trigger annotations. 

Inspired by recent advances in learning sentence classification tasks (e.g., relation extraction and sentiment classification) with explanations or human-written rules~\cite{Li2018GeneralizeSK, Hancock2018TrainingCW, Wang2020Learning, Zhou2019NEROAN}, we propose the concept of an ``entity trigger'' for the task of named entity recognition.
These prior works primarily focused on sentence classification, in which the rules (parsed from natural language explanations) are usually continuous token sequences and there is a single label for each input sentence.
The unique challenge in NER is that we have to deal with rules which are discontinuous token sequences and there may be multiple rules applied at the same time for an input instance.
We address this problem in TMN by jointly learning trigger representations and creating a soft matching module that works in the inference time.

We argue that either dictionary-based distant supervision or active learning can be used in the context of trigger-enhanced NER learning via our framework.
For example, one could create a dictionary using a high-quality corpus and then apply active learning by asking human annotators to annotate the triggers chosen by an active sampling algorithm designed for TMN.
We believe our work sheds light on future research for more cost-effectively using human to learn NER models.

\section{Conclusion}\label{sec:conclusion}

In this paper,
we introduce the concept of ``entity trigger'' as a complementary annotation.
Individual entity annotations provide limited explicit supervision. Entity-trigger annotations add in complementary supervision signals and thus helps the model to learn and generalize more efficiently.
We also crowdsourced triggers on two mainstream datasets and will release them to the community.
We also propose a novel framework TMN which jointly learns trigger representations and soft matching module with self-attention such that can generalize to unseen sentences easily for tagging named entities.  \quad
Future directions with TriggerNER includes: 1) developing models for automatically extracting novel triggers, 2) transferring existing entity triggers to low-resource languages, and 3)  improving trigger modeling with better structured inductive bias (e.g., OpenIE).

\section*{Acknowledgements}
This research is based upon work supported in part by the Office of the Director of National Intelligence (ODNI), Intelligence Advanced Research Projects Activity (IARPA), via Contract No. 2019-19051600007, NSF SMA 18-29268, and Snap research gift. The views and conclusions contained herein are those of the authors and should not be interpreted as necessarily representing the official policies, either expressed or implied, of ODNI, IARPA, or the U.S. Government. We would like to thank all the collaborators in USC INK research lab for their constructive feedback on the work.

\bibliography{akbc} 
\bibliographystyle{acl_natbib}

\end{document}